\pdfoutput=1

\documentclass[11pt]{article}

\usepackage{acl}

\usepackage{times}
\usepackage{latexsym}

\usepackage[T1]{fontenc}

\usepackage[utf8]{inputenc}
\usepackage[inline]{enumitem}

\usepackage{amsmath}
\usepackage{graphicx}
\usepackage{microtype}
\usepackage{booktabs,arydshln}
\usepackage[flushleft]{threeparttable}
\usepackage{tabularx}
\usepackage{multirow}
\usepackage{subcaption}
\usepackage{float}
\usepackage{multicol}

\usepackage[autostyle]{csquotes}

\makeatletter
\def\adl@drawiv#1#2#3{%
        \hskip.5\tabcolsep
        \xleaders#3{#2.5\@tempdimb #1{1}#2.5\@tempdimb}%
                #2\z@ plus1fil minus1fil\relax
        \hskip.5\tabcolsep}
\newcommand{\cdashlinelr}[1]{%
  \noalign{\vskip\aboverulesep
           \global\let\@dashdrawstore\adl@draw
           \global\let\adl@draw\adl@drawiv}
  \cdashline{#1}
  \noalign{\global\let\adl@draw\@dashdrawstore
           \vskip\belowrulesep}}
\makeatother

\usepackage{tcolorbox}
\definecolor{RoyalBlue}{RGB}{0, 114, 188}
\definecolor{Goldenrod}{RGB}{255,222,65}

\usepackage{siunitx}
\newcommand{\mypc}[1]{\hfill(\SI{#1}{\percent})}

\usepackage{makecell}

\title{From Text Segmentation to Smart Chaptering: \\ A Novel Benchmark for Structuring Video Transcriptions}

\author{\textbf{Fabian Retkowski}$^1$, \textbf{Alexander Waibel}$^{1,2}$\\
  $^1$Karlsruhe Institute of Technology, Karlsruhe, Germany\\
  $^2$Carnegie Mellon University, Pittsburgh PA, USA\\
  \texttt{\{retkowski,waibel\}@kit.edu}
}

\begin{document}
\maketitle
\begin{abstract}
Text segmentation is a fundamental task in natural language processing, where documents are split into contiguous sections. However, prior research in this area has been constrained by limited datasets, which are either small in scale, synthesized, or only contain well-structured documents. In this paper, we address these limitations by introducing a novel benchmark \textsc{YTSeg} focusing on spoken content that is inherently more unstructured and both topically and structurally diverse. As part of this work, we introduce an efficient hierarchical segmentation model MiniSeg, that outperforms state-of-the-art baselines. Lastly, we expand the notion of text segmentation to a more practical \enquote{smart chaptering} task that involves the segmentation of unstructured content, the generation of meaningful segment titles, and a potential real-time application of the models.  
\end{abstract}

\section{Introduction}

Text segmentation, also occasionally referred to as document segmentation or topic segmentation, is the task of delimiting the boundaries of topically (or functionally) coherent segments of text, placing them in a hierarchical structure, typically a linear one. Text segmentation has been shown to support a number of applications and downstream tasks where long documents are involved, such as information retrieval \cite{prince_text_2007,shtekh_applying_2018,chivers_ants_2022} or text summarization \cite{zechner_diasumm_2000,cho_toward_2022,liu_end--end_2022,sudmann_current_2023}.

Despite its significance, the field currently lacks robust benchmarks, as it is attested and evident in recent works \cite{lukasik_text_2020, glavas_training_2021}. Most datasets like Choi \cite{choi_advances_2000} either suffer from their small scale or are purely synthetic. In practice, \textsc{Wiki-727K} \cite{koshorek_text_2018} is the only larger-scale available benchmark, consisting of more than 727,000 Wikipedia documents. However, Wikipedia documents may fall short of fully representing the diversity and complexities of real-world text segmentation challenges. These documents are well-structured, informative, and have a fixed style (in accordance with Wikipedia's Manual of Style). This uniformity may not adequately reflect the unstructured and varied nature of text found in other sources.

Structuring a document proves to be particularly valuable in two cases, both of which frequently occur in spoken and conversational content: first, when the content is inherently unstructured; and second, when the content, being structured or semi-structured, lacks explicit or formal organization. Consequently, text segmentation plays a crucial role in many recently rolled-out AI-powered features in applications such as Discord, YouTube, Microsoft Teams, and Zoom. 

Based on this observation and the lack of available, large-scale benchmarks, we have developed a novel benchmark centered around diverse spoken content. We adopt a more holistic and practical approach than prior works, viewing text segmentation as a valuable application that involves both the prediction of segment boundaries as well as the generation of segment titles. Previous works have not considered and evaluated the generation of section headings, which is crucial for practical applications. We term this challenge \textit{smart chaptering} to describe the transformation of unstructured content into a high-level semantic structure with meaningful headings, a critical process for improving document comprehension and organization. This term reflects the additional capabilities required for a system to offer practical utility. Correspondingly, we introduce MiniSeg, a small-scale and state-of-the-art hierarchical segmentation model focused on efficiency, thus viable for use in practical settings. Finally, in addition to traditional offline settings, we also evaluate our approach in online scenarios where real-time processing is crucial, further expanding its practical applicability.

In summary, the contributions of our paper are:

\begin{itemize}
    \item The introduction of a novel larger-scale text segmentation benchmark \textsc{YTSeg} which addresses an important limitation of this research area, the lack of robust benchmarks, and gives researchers a chance to evaluate their models on a benchmark other than \textsc{Wiki-727K}. In addition, it is the first available benchmark around speech text segmentation.
    \item We introduce MiniSeg, an efficient, hierarchical, state-of-the-art text segmentation model that demonstrates the effectiveness of a number of incremental methodological improvements compared to previous models.
    \item We extend the (offline) text segmentation task by online segmentation and title generation. For these tasks, we provide a strong set of baselines on our benchmark.
\end{itemize}

\section{\textsc{YTSeg} Dataset}

As part of this work, we introduce a new benchmark, \textsc{YTSeg}, to evaluate text segmentation systems on less structured and more diverse content than previous benchmarks aimed for. The dataset consists of 19,299 English YouTube videos with their transcripts and chapters. An example of a YouTube video organized into chapters can be found in the screenshot provided in Figure \ref{fig:screenshot}. We processed the data to adapt it for the text segmentation task, aligning the sentences in the transcription with video chapters. In addition, we release \textsc{YTSeg[Titles]} for the training and evaluation of generative models predicting the chapter titles. By including video and audio data, the benchmark also paves the way for multi-modal approaches. The dataset, along with the instructions and scripts, is available online and released under a CC BY-NC-SA 4.0 license\footnote{\url{https://huggingface.co/datasets/retkowski/ytseg}}.

\begin{table*}[t]
\centering
  \begin{subtable}[t]{0.4\textwidth}
   \centering
    \begin{tabular}{>{\raggedright\arraybackslash}p{3cm}r}
        \toprule
        \textbf{Channel Name} & \textbf{\# Videos} \\
        \midrule
        YaleCourses & 1015 \\
        The List & 984 \\
        Mashed & 904 \\
        Bestie & 577 \\
        Google Cloud Tech & 551 \\
        Linus Tech Tips & 461 \\
        Unveiled & 425 \\
        Flipping Physics & 390 \\
        Looper & 385 \\
        GeologyHub & 364 \\
        \bottomrule
    \end{tabular}
    \caption{YouTube Channels by Number of Videos}
    \label{tab:yt_top_count}
\end{subtable}
  \quad
  \begin{subtable}[t]{0.4\textwidth}
    \centering
    \begin{tabular}{>{\raggedright\arraybackslash}p{3cm}r}
        \toprule
        \textbf{Channel Name} & \textbf{Length[h]} \\
        \midrule
        YaleCourses & 907.05 \\
        CS50 & 276.21 \\
        Rich Roll & 179.16 \\
        Andrew Huberman & 178.92 \\
        The List & 168.27 \\
        Mashed & 159.24 \\
        Tech With Tim & 131.37 \\
        Linus Tech Tips & 120.14 \\
        SAS Users & 106.12 \\
        David Bombal & 93.09 \\

        \bottomrule
    \end{tabular}
    \caption{YouTube Channels by Content Length}
    \label{tab:yt_top_length}
\end{subtable}
\caption{YouTube Channels by Number of Videos and Content Length}
\label{tab:yt_top}
\end{table*}

\subsection{Collection}

For the dataset, we utilize \texttt{yt-dlp} to collect videos, their transcripts, and their chapters. The dataset collection is limited to videos with closed captions and chapters provided by the content creator. In both cases, YouTube exposes whether the information is automatically generated.\footnote{While for closed captions, this information can be accessed directly via \texttt{yt-dlp}, for chapters, it is derived from raw data returned by \texttt{yt-dlp}'s low-level APIs.} We note that only a small subset of YouTube videos fall under this category, which motivates the following procedure.

In the first step, we define a wide variety of seed keywords as listed in Section \ref{sec:seed}. Their purpose is to surface a higher-quality and diverse set of videos that are more likely to have manually provided closed captions and chapters. Then, we utilize the YouTube search and various search filters (e.g., to filter videos without closed captions or to surface more recent content or long-form videos). Based on the video search results, we select corresponding channels to be crawled after reviewing a sample of the channel's videos for the audio language and the quality of its closed captions and chapters.

\subsection{Preprocessing}
\label{section:preprocessing}

Following previous benchmarks for the text segmentation task, our dataset aims to provide segment boundaries on a sentence level. For this, we sentence tokenize the closed captions using the pre-trained \textsc{Punkt} tokenizer available in the NLTK library \cite{bird_natural_2009}. We annotate the sentences with respective timestamps based on the closed captions. It is important to note that the closed captions do not always respect sentence boundaries, which may necessitate potential splitting and joining of sentences while linearly interpolating timestamps based on their character length. Unlike purely textual datasets, the segment boundaries might not necessarily agree with sentence boundaries (i.e., YouTube chapters can start or end in the middle of a sentence). Thus, sentences spanning two chapters are assigned to the chapter with the greater time-based overlap. We also observed instances where sentences remain unassigned when the first chapter starts later or the last chapter ends earlier than the first, respectively, the last caption. To address this issue, we add an additional \enquote{Intro} or \enquote{Outro} chapter in these cases.

We exclude all videos for which inconsistent timestamps cannot be fixed\footnote{In a number of instances, we observed certain inconsistencies in the provided timestamps fixable by simple rules.}, final sanity checks are not passed, or our procedure returns an error. These errors can stem from various reasons, such as empty captions, transcripts without punctuation, or a malformed VTT format. This affects $4.70\%$ of the collected videos.

Finally, the data is split into stratified partitions for training, validation, and testing based on the channel identifier (see Table \ref{table:ytseg_split}). As part of this process, channels with only a single video form a separate group.

\subsection{Data Statistics}

\begin{figure}[h]
\centering
\includegraphics[width=0.5\textwidth,clip]{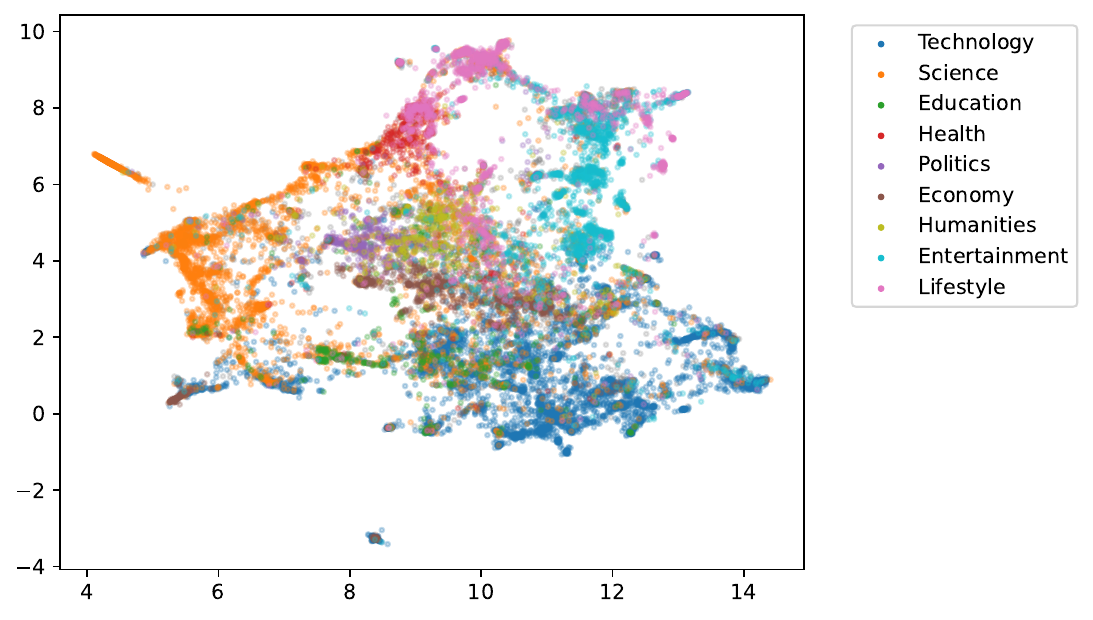}
\caption{UMAP \cite{mcinnes_umap_2018} plot of \textsc{YTSeg} video titles, embedded using Instructor \cite{su_one_2023}. Category labels are assigned through zero-shot classification with LLaMA 2 \cite{touvron_llama_2023}.}
\label{fig:domain_analysis}
\end{figure}

The dataset comprises 19,299 videos from 393 channels, amounting to 6,533 content hours. The topics are wide-ranging, covering domains such as science, lifestyle, politics, health, economy, and technology. The videos are from various types of content formats, such as podcasts, lectures, news, corporate events \& promotional content, and, more broadly, videos from individual content creators. Table \ref{tab:yt_top_count} and the analysis depicted in Figure \ref{fig:domain_analysis} offer insights into the dataset's diversity, while Table \ref{tab:yt_top_length} shows that the content hours are dominated by the long-form formats such as podcasts and lectures. The dataset's structural diversity is also evident in its data statistics, as depicted in Table \ref{tab:data-stats}. In contrast to \textsc{Wiki-727K}, our benchmark exhibits a greater number of segments per document and a higher number of sentences per segment while simultaneously showing a wider variation.

\begin{table}[h]
    \centering
    \tabcolsep=0.100cm
    \small
    \begin{threeparttable}
    
    \begin{tabular}{>{\itshape}l *{3}{c}}
        \toprule
        & \textbf{\textsc{YTSeg}} & \textbf{\textsc{Wiki-727K}} & \\
        \midrule
        Document Length [Sent.] & 196.2 $\pm$\ 267.2 & 57.6 $\pm$\ 46.9\\ 
        Video Length [min.] & 20.3 $\pm$\ 25.3 & --\\ 
        \cdashlinelr{1-3}
        Segment Length [Sent.] & 21.5 $\pm$\ 34.2 & 13.6 $\pm$\ 20.3\\ 
        Segment Duration [min.] & 2.49 $\pm$\ 2.98 & -- \\ 
        Segments per Document & 9.12 $\pm$\ 5.42 & 3.48 $\pm$\ 2.23\\ 
        \cdashlinelr{1-3}
        Title Length [Words]  & 4.03 $\pm$\ 2.75 & 2.01 $\pm$\ 1.49 \\
        Concentration Index\tnote{1} & 9.50\% & 24.96\% \\
        \bottomrule
    \end{tabular}

    \begin{tablenotes}[flushleft]\footnotesize\parindent=1em
    \setlength{\columnsep}{0.8cm}
    \setlength{\multicolsep}{0cm}
    \item[1]with $n=20$
    \end{tablenotes}
    \end{threeparttable}
    
    \caption{Data Statistics for \textsc{YTSeg} and \textsc{Wiki-727K}}
    \label{tab:data-stats}
\end{table}

\subsection{Chapter Titles}
Based on the same data partitioning outlined in Section \ref{section:preprocessing}, we prepared another view on the same dataset, providing 173,195 pairs of sections and chapter titles. We refer to this dataset view as \mbox{\textsc{YTSeg[Titles]}}. As a result of the same data partitioning, all section title pairs of a particular video will be assigned to the same data partition. The resulting data split can be found in Table \ref{table:ytsegtitle_split}. We removed every pair for which the section title exceeds 75 characters to exclude atypically and excessively lengthy titles\footnote{Some of these lengthy titles tend to be complete sentences or summaries, deviating from our understanding of a title. We also note the computational advantages of excluding them.} accounting for 1.5\% of the total number of titles. Despite this, the title length is meaningfully longer and more diverse than in titles in the \textsc{Wiki-727K} dataset (see Table \ref{tab:data-stats}). We also point out that the titles in the \textsc{Wiki-727K} dataset are highly concentrated. The 20 most frequent titles in the dataset account for 24.96\% of the overall dataset, with the title \enquote{History} alone constituting 7.53\% of it. On the contrary, for the \textsc{YTSeg} dataset, this concentration is notably lower, as the top 20 titles collectively represent just 9.50\% of the dataset. These titles predominantly consist of functional segments like \enquote{Introduction} or \enquote{Conclusion}.

\section{Methodology}

In the following, we present the models designed for the text segmentation and title generation task, which are used in our experiments and applied to the newly introduced benchmark. We also elaborate on the modifications we have made to adapt them for online implementation.

\begin{figure}[h]
\centering
\includegraphics[width=0.5\textwidth,trim={1cm 0.5cm 0 1.25cm},clip]{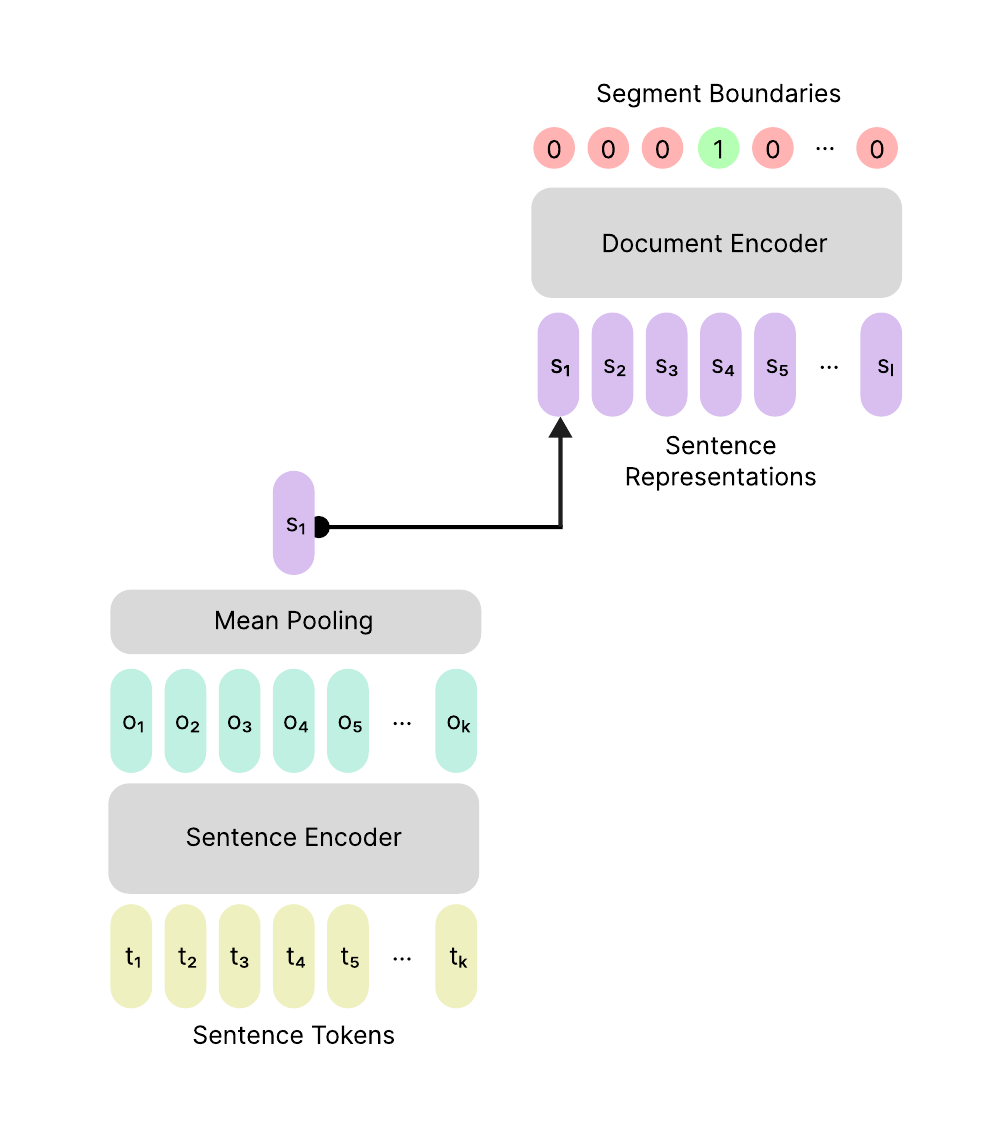}
\caption{The hierarchical architecture of the segmentation model consists of a sentence encoder and a document encoder returning the binary segment boundaries.}
\label{fig:architecture}
\end{figure}

\begin{figure*}[t]
\centering
\includegraphics[width=1\textwidth,trim={0.5cm 0.7cm 1.2cm 1.5cm},clip]{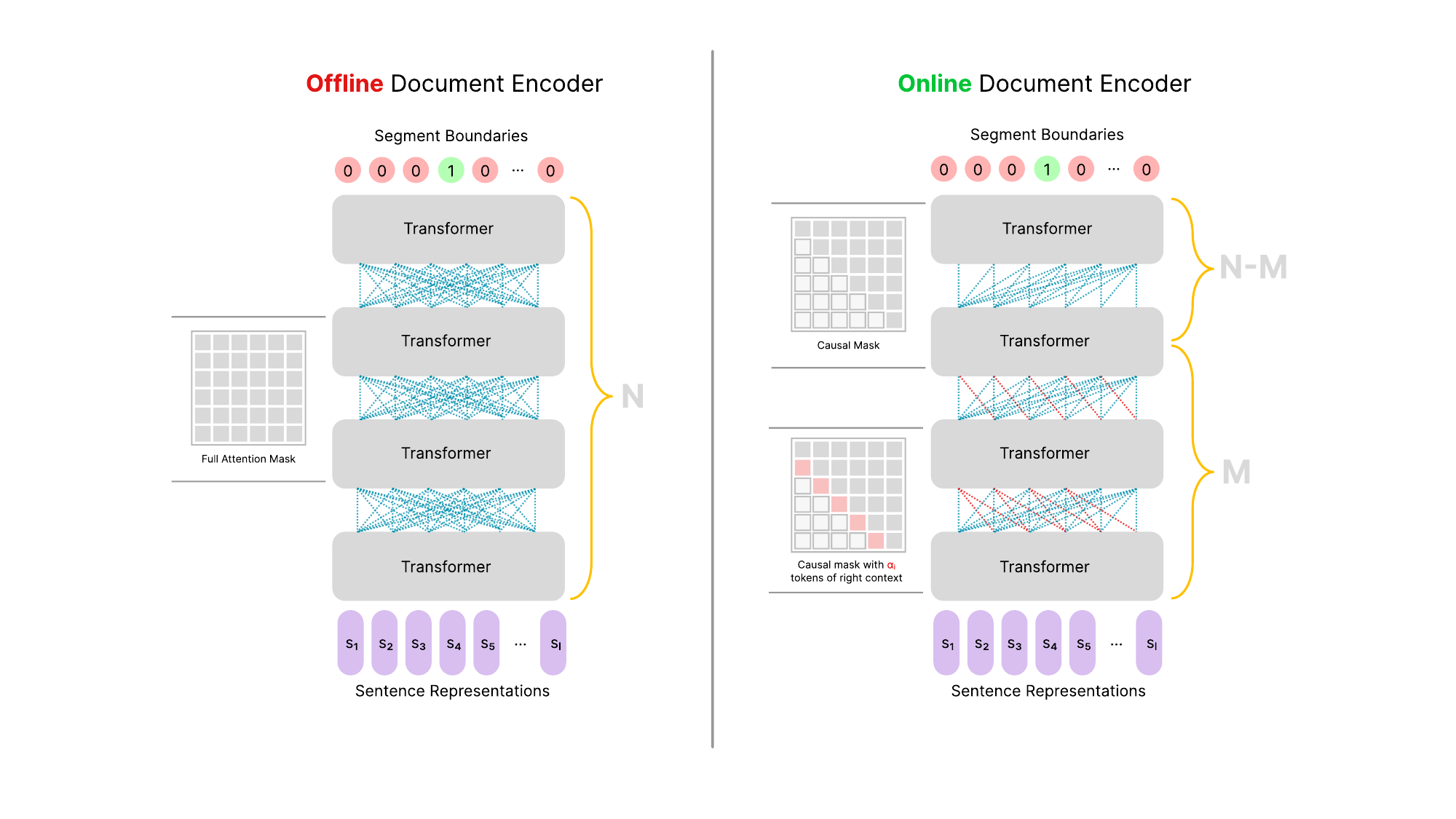}
\caption{Our offline document encoder is a typical transformer encoder with $N$ transformer layers, each of which applies a full attention mask. Consequently, the encoder can attend to the whole document. In contrast, our online document encoder has $N-M$ layers with causal attention masks that only allow attention to past context, while the initial $M$ layers have attention masks with limited right-side context, that, over these $M$ layers, accumulate to a defined future context size $c$.}
\label{fig:document_encoder}
\end{figure*}

\subsection{Offline Segmentation}
\label{sec:off_seg}
The model used in our experiments (see the architecture outlined in Figure \ref{fig:architecture}) closely resembles the previous work of 
\citet{koshorek_text_2018} and \citet{lukasik_text_2020}. Both proposed a hierarchically structured network consisting of a sentence encoder and a document encoder. The sentence encoder processes each token within a sentence to generate a corresponding sentence representation. Following this, the document encoder performs a sequence labeling task, where its aim is to predict whether each sentence serves as a segment boundary. The network is trained in a supervised fashion using a binary classification objective.

In the following, we highlight the methodological differences between our MiniSeg model and the hierarchical BERT architecture outlined in \citet{lukasik_text_2020}.

\begin{itemize}
    \item We utilize a pre-trained sentence transformer based on MiniLM \cite{reimers_sentence-bert_2019, wang_minilm_2020} for the sentence encoder (33M parameters). This network has specifically been trained on paraphrase data in a siamese network structure to produce meaningful sentence representations.
    \item For the document encoder, we use a randomly initialized RoFormer encoder with 12 layers, 8 attention heads, and  384-dimensional embeddings (26M parameters). This Transformer variant uses rotary positional embeddings (RoPE) introduced by \citet{su_roformer_2021}.
    \item Motivated by the class imbalance, we opted to use a weighted binary cross-entropy term. In a thoughtful adjustment, we assigned double the weight to segment boundaries with $w=[1,2]$. A similar reweighting has been performed by \citet{ghosh_topic_2022}.
    \item Instead of using the \texttt{[CLS]} token, we apply mean pooling on the output embeddings to create fixed-sized sentence representations, as it has been shown to outperform other pooling strategies \cite{reimers_sentence-bert_2019}.
    \item While training, we randomly sample a subset of gradients for the sentence-encoding sub-network in each backward pass. We set this gradient sampling rate to $0.5$, meaning half of the documents are backpropagated through the sentence encoder. In our experiments, this has been shown to reduce the memory requirement while having a regularizing effect and improving the final performance.
\end{itemize}

\subsection{Online Segmentation}

Our proposed online segmentation model mirrors the offline segmentation model in its architecture (see Section \ref{sec:off_seg}). The major difference is that we limit the future context that the model can process, for which we use a different masking strategy that we refer to as \textit{progressive context accumulation}. Starting from a purely causal model, we replace a subset of causal attention masks with masks that allow attention to a controlled amount of future context. The corresponding architecture and idea are illustrated in Figure \ref{fig:document_encoder}.

Specifically, in the early $M$ layers of the document encoder, we employ causal attention masks with an offset that is the limited right-side context $\alpha_i$ where $i\in[1,M]$ is the index of the layer. These masks provide each layer with selective access to a portion of additional future context, and their sizes sum up to our predefined target future context size, denoted as $c$. In the later $N-M$ layers, we transition back to causal attention masks to prevent any additional future context from leaking into the predictions.

This approach introduces a structural hyperparameter $\alpha$ defining the partitioning distribution of the total future context size $c$ to be allocated to each of the first $M$ layers. We note the relation between the introduced hyperparameters: $M=|\alpha|$ and $c=\sum{\alpha_i}$.

\subsection{Title Generation}
We fine-tune a BART-large \cite{lewis_bart_2020} model on our \mbox{\textsc{YTSeg[Titles]}} dataset for section title generation. BART is a transformer encoder-decoder model pre-trained on a denoising task.

\begin{itemize}
    \item For \textbf{online title generation}, we limit the amount of text we provide from a given section to the model for generating titles. The term \textit{input span} further refers to the number of starting sentences from a section. This approach enables titles to be prematurely generated in an online setting while maintaining a defined latency (in terms of sentences).
    \item \textbf{Conditional title generation}: To incorporate the context of the document's structure, we prepend previous section titles. This way, the model is conditioned on both the content of the current section and the preceding context, allowing the generation of more relevant and coherent titles. We note that this approach requires generating titles sequentially, which affects offline title generation. In contrast, for online title generation, titles are always generated sequentially.
\end{itemize}

\section{Experiments}

\subsection{Segmentation}

\begin{table*}[t]
\small
\tabcolsep=0.09cm
\centering
\begin{threeparttable}
\centering
\begin{tabular}{llccccccccccc}
\toprule
      &              & P (↑)        & R (↑)        & F1 (↑)     & P\textsubscript{$k$} (↓) & B\tnote{1} (↑) \\
    \midrule
    \multirow{4}{*}{\footnotesize\textsc{Wiki-727K}}
    & Bi-LSTM\tnote{2,4}         & 69.3\hphantom{0} $\pm$\ 0.1\hphantom{0}   & 49.5\hphantom{0} $\pm$\  0.2\hphantom{0} & 57.7\hphantom{0} $\pm$\ 0.1\hphantom{0}  & -- & -- \\
    & CS BERT\tnote{3,4}          & 69.1\hphantom{0} $\pm$\ 0.1\hphantom{0} & 63.2\hphantom{0} $\pm$\ 0.2\hphantom{0} & 66.0\hphantom{0} $\pm$\ 0.1\hphantom{0} & -- & -- \\
    & Hier. BERT\tnote{3,4}          & \textbf{69.8\hphantom{0} $\pm$\ 0.1\hphantom{0}} & 63.5\hphantom{0} $\pm$\ 0.1\hphantom{0} & 66.5\hphantom{0} $\pm$\ 0.1\hphantom{0} & -- & -- \\
    & MiniSeg (Ours)         & 68.57 $\pm$\ 0.13  & \textbf{70.76 $\pm$\ 0.13} &  \textbf{69.65 $\pm$\ 0.09} & 17.57 $\pm$\ 0.06 & 59.81 $\pm$\ 0.12 \\
    \midrule
    \midrule
    \multirow{1}{*}{\footnotesize\textsc{YTSeg}}
    & MiniSeg         & 45.44 $\pm$\ 0.83 & 41.48 $\pm$\ 0.85 & 43.37 $\pm$\ 0.60 & 28.73 $\pm$\ 0.39 & 35.74 $\pm$\ 0.68 \\
    \midrule
    \multirow{8}{*}{\footnotesize\textsc{Wiki-727K} $\rightarrow$ \textsc{YTSeg}}
    & MiniSeg        & 48.30 $\pm$\ 0.84 & 43.56 $\pm$\ 0.84 & 45.81 $\pm$\ 0.60 & 27.13 $\pm$\ 0.43 & 37.89 $\pm$\ 0.70 \\
    \cdashlinelr{2-7}
    & MiniSeg ($c=0$)         & 43.69 $\pm$\ 0.79 & 37.49 $\pm$\ 0.76 & 40.35 $\pm$\ 0.55 & 29.81 $\pm$\ 0.38 & 33.11 $\pm$\ 0.72 \\
    & MiniSeg ($c=1$)         & 45.05 $\pm$\ 0.82 & 40.05 $\pm$\ 0.80 & 42.41 $\pm$\ 0.58 & 28.70 $\pm$\ 0.40 & 34.72 $\pm$\ 0.79 \\
    & MiniSeg ($c=3$)         & 46.02 $\pm$\ 0.88 & 41.45 $\pm$\ 0.77 & 43.61 $\pm$\ 0.58 & 28.08 $\pm$\ 0.43 & 36.13 $\pm$\ 0.76 \\
    & MiniSeg ($c=5$)         & 46.24 $\pm$\ 0.75 & 42.23 $\pm$\  0.91 & 44.15 $\pm$\ 0.60 & 27.91 $\pm$\ 0.80 & 36.62 $\pm$\ 0.78 \\
    & MiniSeg ($c=8$)         & 46.92 $\pm$\ 0.80 & 41.89 $\pm$\ 0.79 & 44.26 $\pm$\ 0.57 & 27.68 $\pm$\ 0.42 & 36.81 $\pm$\ 0.77 \\
    & MiniSeg ($c=10$)         & 45.99 $\pm$\ 0.81 & 41.31 $\pm$\ 0.89 & 43.52 $\pm$\ 0.61 & 27.97 $\pm$\ 0.39 & 36.35 $\pm$\ 0.74 \\
    & MiniSeg ($c=20$)         & 46.46 $\pm$\ 0.89 & 42.34 $\pm$\ 0.83 & 44.30 $\pm$\ 0.61 & 27.95 $\pm$\ 0.41 & 37.07 $\pm$\ 0.76 \\
\bottomrule
\end{tabular}

    \begin{tablenotes}[flushleft]\footnotesize\parindent=1em
    \setlength{\columnsep}{0.8cm}
    \setlength{\multicolsep}{0cm}
    \begin{multicols}{2}
    \item[1]Boundary Similarity \cite{fournier_evaluating_2013}
    \item[2] \citet{koshorek_text_2018}
    \item[3] \citet{lukasik_text_2020}
    \item[4] Results as reported by \citet{lukasik_text_2020}.
    \end{multicols}
    \end{tablenotes}
\end{threeparttable}
\caption{Results of our text segmentation models and baselines on \textsc{Wiki-727K} and \textsc{YTSeg}. Standard deviations are estimated by bootstrapping the test set 100 times, similar as in \citet{lukasik_text_2020}.}
\label{tab:results_seg}
\end{table*}

\begin{table}[t]
\scriptsize
\tabcolsep=0.09cm
\centering
\begin{tabular}{llccccccccccc}
\toprule
    & & P (↑) & R (↑) & F1 (↑) \\
    \midrule
    \multicolumn{5}{c}{\textbf{Zero-Shot}} \\
    \cdashlinelr{1-5}
    & \textsc{Wiki-727K} & \hphantom{0}1.58 $\pm$\ 0.89 & 15.85 $\pm$\ 3.61 & \hphantom{0}2.87 $\pm$\ 1.47 \\
    & \textsc{YTSeg} & 12.55 $\pm$\ 4.21 & \hphantom{0}8.30 $\pm$\ 2.78 & \hphantom{0}9.99 $\pm$\ 2.41 \\
    & \textsc{Wiki-727K} $\rightarrow$ \textsc{YTSeg} & \hphantom{0}6.09 $\pm$\ 1.92 & \textbf{18.07 $\pm$\ 3.36} & \hphantom{0}9.11 $\pm$\ 2.19 \\
    \midrule
    \multicolumn{5}{c}{\textbf{Fine-Tuned on \textsc{QMSum}}} \\
    \cdashlinelr{1-5}
    & No Pre-Training & 21.62 $\pm$\ 4.54 & 15.61 $\pm$\ 3.58 & 18.13 $\pm$\ 2.90 \\
    & \textsc{Wiki-727K} & 23.45 $\pm$\ 5.42 & 11.88 $\pm$\  2.75 & 15.77 $\pm$\ 2.71 \\
    & \textsc{YTSeg} & \textbf{31.09 $\pm$\ 5.35} & 16.92 $\pm$\ 3.20 & \textbf{21.92 $\pm$\ 2.99} \\
    & \textsc{Wiki-727K} $\rightarrow$ \textsc{YTSeg} & 25.21 $\pm$\ 4.71 & 15.82 $\pm$\ 3.15 & 19.44 $\pm$\ 2.76 \\
\bottomrule
\end{tabular}
\caption{Text segmentation results of MiniSeg on the \textsc{QMSum} dataset, both zero-shot and fine-tuned.}
\label{tab:results_qmsum}
\end{table}

\begin{table*}[t]
\centering
\begin{threeparttable}
\small
\centering
\begin{tabular}{llccccccccccc}
\toprule
      &              & P (↑)        & R (↑)        & F1 (↑)     & P\textsubscript{$k$} (↓) & B (↑) \\
    \midrule
   & MiniSeg         & 45.44 $\pm$\ 0.83 & 41.48 $\pm$\ 0.85 & 43.37 $\pm$\ 0.60 & \textbf{28.73 $\pm$\ 0.39} & \textbf{35.74 $\pm$\ 0.68} \\
   \cdashlinelr{1-7}
   & \hspace*{0.3cm} w/o WBCE        & \textbf{48.76 $\pm$\ 0.95} & 31.66 $\pm$\ 0.77 & 38.39 $\pm$\ 0.64 & 30.53 $\pm$\ 0.36 & 30.42 $\pm$\ 0.71 \\
   & \hspace*{0.3cm} w/o RoPE\tnote{1}        & 42.13 $\pm$\ 0.70 & \textbf{42.05 $\pm$\ 0.81} & 42.09 $\pm$\ 0.54 & 30.59 $\pm$\ 0.47 & 33.75 $\pm$\ 0.69 \\
   & \hspace*{0.3cm} w/o Pre-Training\tnote{2}  & 38.73 $\pm$\ 0.81 & 28.62 $\pm$\ 0.61 & 32.92 $\pm$\ 0.50 & 33.52 $\pm$\ 0.36 & 25.45 $\pm$\ 0.59 \\
   & \hspace*{0.3cm} with \texttt{[CLS]} pooling  & 45.76 $\pm$\ 0.80 & 41.39 $\pm$\ 0.85 & \textbf{43.47 $\pm$\ 0.59} & 28.87 $\pm$\ 0.39 & 35.54 $\pm$\ 0.68 \\
   & \hspace*{0.3cm} w/o gradient sampling  & 43.41 $\pm$\ 0.83 & 40.38 $\pm$\ 0.79 & 41.84 $\pm$\ 0.57 & 29.94 $\pm$\ 0.41 & 34.10 $\pm$\ 0.72 \\

\bottomrule
\end{tabular}
    \begin{tablenotes}[flushleft]\footnotesize\parindent=1em
    \setlength{\columnsep}{0.8cm}
    \setlength{\multicolsep}{0cm}
    \item[1]A standard transformer with sinusoidal positional encodings is used as the document encoder.
    \item[2]The weights of MiniLM, the sentence transformer, are initialized randomly. 
    \end{tablenotes}
\end{threeparttable}
\caption{Results of ablated versions of MiniSeg on the \textsc{YTSeg} dataset.}
\label{tab:ablation_results}
\end{table*}

We perform the following experiments to evaluate our benchmark and our segmentation model:
\begin{itemize}
    \item First, we train our introduced MiniSeg model on the established benchmark \textsc{Wiki-727K} and compare it against the baselines of \citet{koshorek_text_2018} and \citet{lukasik_text_2020}. For a fair comparison, we use the same setup as in \citet{koshorek_text_2018} by predicting top-level sections of the document and using the original preprocessing scripts. 
    \item We establish a benchmark for the \textsc{YTSeg} dataset employing our MiniSeg model. This involves training the model on the dataset to set the baseline performance.
    \item In addition, we experiment with a two-stage training process where we first do a task adaptational pre-training of the model on the \textsc{Wiki-727K} dataset and then fine-tune it on our \textsc{YTSeg} benchmark.
    \item We test and fine-tune our model on \textsc{QMSum} \cite{zhong_qmsum_2021} to evaluate whether our dataset and model can improve the segmentation of even more unstructured content such as meetings. This dataset provides 232 segmented meetings.
    \item Finally, we train online segmentation models with different future context sizes $c$. The corresponding partitioning $\alpha$ for each setting can be found in Table \ref{tab:partition}.
\end{itemize}

We evaluate our segmentation models using a combination of standard binary classification metrics, such as precision, recall, and F1 score, as well as metrics specifically tailored for text segmentation tasks, including P\textsubscript{$k$} as introduced in the work of \citet{beeferman_statistical_1999} and Boundary Similarity, as discussed in \citet{fournier_evaluating_2013}. The results of the experiments are presented in Table \ref{tab:results_seg} and \ref{tab:results_qmsum}.

\paragraph{MiniSeg.} The experiments presented in Table \ref{tab:results_seg} demonstrate that MiniSeg outperforms the baselines, namely Bi-LSTM, cross-segmenter BERT, and hierarchical BERT, on the established \textsc{Wiki-727K} benchmark, even though it is equipped with only 59 million parameters. Given this parameter count, it is meaningfully more efficient compared to state-of-the-art baselines such as hierarchical BERT (220 million parameters) and cross-segmenter BERT (336 million parameters). Several factors contribute to the observed performance as demonstrated in an ablation study shown in Table \ref{tab:ablation_results}, with the strength of the pre-trained sentence encoder emerging as one of the most crucial contributors. It is worth noting that BERT \cite{devlin_bert_2019}, as used by \citet{lukasik_text_2020}, was not trained to represent sentences in particular and that \citet{liu_roberta_2019} have described BERT as \enquote{significantly undertrained}. Additionally, our approach relies on a weighted cross-entropy loss function, allowing us to balance precision and recall. Importantly, while our model exhibits lower precision compared to the baselines and ablated version, it excels in terms of recall. Smaller incremental gains can be attributed to RoPE and the sampling of gradients. Lastly, no noticeable effect is observed when replacing \texttt{[CLS]} pooling with mean pooling.

\paragraph{Task Adaptation.} We find that the task-adaptational pre-training with \textsc{Wiki-727K} improves the result on the \textsc{YTSeg} benchmark (see Table \ref{tab:results_seg}). This outcome contrasts with the findings of \citet{ghosh_topic_2022}, who reported that such a pre-training step has a negative or negligible effect when applied to semi-structured content. While the domain is different (YouTube videos versus chat conversations), we emphasize that the dataset in \citet{ghosh_topic_2022} is synthetically constructed and, as such, is qualitatively different from \textsc{Wiki-727K} and \textsc{YTSeg} presumably contributing to the varying effectiveness of the pre-training.

\paragraph{Meeting Segmentation.} Our experiments on \textsc{QMSum} displayed in Table \ref{tab:results_qmsum} reveal that the pre-training with \textsc{YTSeg} improves the final performance of models on \textsc{QMSum}. Even in zero-shot conditions, the beneficial effect of \textsc{YTSeg} becomes apparent. This underscores the domain proximity of video content and meetings, both of which are less structured and are spoken in nature. However, we note that due to its small size, \textsc{QMSum} is not a robust benchmark, and the effect of \textsc{Wiki-727K} remains inconclusive.

\paragraph{Online Segmentation.} In the results (see Table \ref{tab:results_seg}), we observe noticeable jumps in performance when increasing the future context size. However, diminishing returns set in after about three to five sentences of future context, especially when considering the latency trade-off in online segmentation. This strongly suggests that local context is more important. In fact, even a model without future context at all scores solidly (with the performance only increasing from 40.35 to 45.81 for a model without any future context and one with global context).

\subsection{Title Generation}

\begin{table*}[t]
\footnotesize
\centering
\begin{threeparttable}
\centering
    \begin{tabular}{llccccccccccc}
    \toprule
                              &              & R1 (↑)        & R2 (↑)        & RL (↑)     & BS\tnote{1} (↑) \\
                            \midrule
                            \multirow{5}{*}{No Context}
                            & BART         & 36.42 $\pm$\ 0.36  & 17.03 $\pm$\ 0.28 & 36.19 $\pm$\ 0.35 & -4.21 $\pm$\ 0.02 \\
                            \cdashlinelr{2-6}
                            & BART ($s=1$) & 25.02 $\pm$\ 0.31  & 10.98 $\pm$\ 0.23  & 24.87 $\pm$\ 0.30 & -4.97 $\pm$\ 0.02  \\
                            & BART ($s=3$) & 31.40 $\pm$\ 0.31 & 14.32 $\pm$\ 0.26  & 31.21 $\pm$\ 0.32 & -4.51 $\pm$\ 0.02 \\
                            & BART ($s=5$) & 33.64 $\pm$\ 0.31  & 15.61 $\pm$\ 0.29 & 33.42 $\pm$\  0.35 & -4.37 $\pm$\ 0.02  \\
                            & BART ($s=10$)& 34.83 $\pm$\ 0.33  & 16.18 $\pm$\ 0.27 & 34.60 $\pm$\  0.37 & -4.30 $\pm$\ 0.02  \\
                            \midrule
                            \multirow{5}{*}{Previous Titles}
                            & BART         & 42.79 $\pm$\ 0.34 & 22.07 $\pm$\ 0.30 & 42.45 $\pm$\ 0.31 & -3.87 $\pm$\ 0.02 \\
                            \cdashlinelr{2-6}
                            & BART ($s=1$) & 27.66 $\pm$\ 0.35  & 11.87 $\pm$\ 0.26   & 27.47 $\pm$\ 0.29 & -4.83 $\pm$\ 0.02  \\
                            & BART ($s=3$) & 36.33 $\pm$\ 0.37  & 17.74 $\pm$\ 0.31   & 36.08 $\pm$\ 0.30 & -4.25 $\pm$\ 0.02  \\
                            & BART ($s=5$) & 36.02 $\pm$\ 0.32  & 19.68 $\pm$\ 0.32   & 35.74 $\pm$\ 0.29 & -4.09 $\pm$\ 0.02   \\
                            & BART ($s=10$)& 41.52 $\pm$\ 0.33  & 21.24 $\pm$\ 0.30   & 41.21 $\pm$\ 0.30 & -3.94 $\pm$\ 0.03   \\
    
    \bottomrule
\end{tabular}
    \begin{tablenotes}[flushleft]\footnotesize\parindent=1em
    \item[1]BARTScore \cite{yuan_bartscore_2021}
    \end{tablenotes}
\end{threeparttable}
\caption{Results of the title generation models on the \textsc{YTSeg[Titles]} dataset.}
\label{tab:title_gen}
\end{table*}

\begin{figure*}[t]
    \centering
    {
    \begin{tcolorbox}[colframe=RoyalBlue, colback=white]
    \small{\textbf{Configuring Cloud Operations on Google Cloud -- Google Cloud Tech (\href{https://www.youtube.com/watch?v=xIaaGef1QvI}{\texttt{\mbox{xIaaGef1QvI})}}}}
    \\[5px]
    \scriptsize
    \renewcommand{\arraystretch}{1.15}
\begin{tabular}{@{} p{0.32\linewidth} !{\color{RoyalBlue}\vrule width 0.5pt} p{0.32\linewidth} !{\color{RoyalBlue}\vrule width 0.5pt} p{0.32\linewidth} @{}}
        \colorbox{Goldenrod}{Intro} & Intro  & Introduction \& agenda \\
        \colorbox{Goldenrod}{Intro} & How to know what’s going on in the cloud & How the operations components play together \\
        Operational use cases & Exploring the products of interest for the operations team & Google Cloud Operations Suite \\
        Site Reliability Engineering & Google SRE & SRE Practices \\
        Customer success story: Krikey & Customer success story: Krikey & Customer Story - Krikey \\
        Wrap-up & Wrap up & Wrap up \& additional resources \\[0.15cm]
        \hspace*{\fill}\textcolor{gray}{\small{\textbf{Generated (No Context)}}}\hspace*{\fill} & \hspace*{\fill}\textcolor{gray}{\small{\textbf{Generated (Previous Titles)}}}\hspace*{\fill} & \hspace*{\fill}\textcolor{gray}{\small{\textbf{Reference}}}\hspace*{\fill}
    \end{tabular}
    \end{tcolorbox}
    \vspace{-0.4em}
    \subcaption{An exemplary output showing duplicate section titles.}
    }
    {
    \begin{tcolorbox}[colframe=RoyalBlue, colback=white]
    \small{\textbf{8 Email Etiquette Tips - How to Write Better Emails at Work -- Harvard Business Review (\href{https://www.youtube.com/watch?v=1XctnF7C74s}{\texttt{1XctnF7C74s})}}}
    \\[5px]
    \scriptsize
    \renewcommand{\arraystretch}{1.15}
\begin{tabular}{@{} p{0.32\linewidth} !{\color{RoyalBlue}\vrule width 0.5pt} p{0.32\linewidth} !{\color{RoyalBlue}\vrule width 0.5pt} p{0.32\linewidth} @{}}
        Intro & Intro & Why bother with email etiquette? \\
        \colorbox{Goldenrod}{Step 1:} Have a call to action if appropriate & 1. Have a call to action when appropriate & Include CTA in subject line \\
        Stick with one email thread for the same topic & 2. Stick with one email thread for the same topic & One email thread per topic \\
        \colorbox{Goldenrod}{Tip \#3} & 3. Explain why you added in or took out recipients & Manage recipients \\
        \colorbox{Goldenrod}{Tip \#5:} summarize the sender’s main points & 4. Include your main point first followed by context & Summarize in your reply \\
        \colorbox{Goldenrod}{Tip \#2:} Include the context & 5. summarize the sender's main points in your reply & Start with the main point \\
        \colorbox{Goldenrod}{6.} Hyperlink Whatever Possible & 6. Hyperlink whatever possible & Hyperlink whenever possible \\
        Change Default Setting to reply instead of reply all & 7. Change default setting to reply instead of reply all & Change default setting to "Reply" (not "Reply all") \\
        Change Undo Send Option to 30 Seconds & 8. Change undo send option to 30 seconds & Change undo send options \\
        Outro & Outro & Outro \\[0.15cm]
        \hspace*{\fill}\textcolor{gray}{\small{\textbf{Generated (No Context)}}}\hspace*{\fill} & \hspace*{\fill}\textcolor{gray}{\small{\textbf{Generated (Previous Titles)}}}\hspace*{\fill} & \hspace*{\fill}\textcolor{gray}{\small{\textbf{Reference}}}\hspace*{\fill}
    \end{tabular}
    \end{tcolorbox}
    \vspace{-0.4em}
    \subcaption{An exemplary output showing inconsistent numbering and formatting.}
    }
 \caption{A comparison of chapter titles generated by the fine-tuned BART models, both without any context and with previously generated titles, and the reference titles on select examples from \textsc{YTSeg}'s validation dataset. A lack of context, as \colorbox{Goldenrod}{highlighted}, leads to repetitive titles, coherence gaps, and variations in writing styles.}
\label{fig:titles_examples}
\end{figure*}

We conducted a series of experiments to evaluate the performance of our title generation model, summarized as follows:
\begin{itemize}
    \item In our initial experiment, we conducted a comparative analysis of fine-tuning the BART model under two distinct conditions. The first involved training the model to generate titles solely based on the current section text, devoid of any contextual information. In the second setting, the model was trained with the added context of previous titles.
    \item For the online setting, we assessed different scenarios where the model receives a limited number of starting sentences $s\in[1,3,5,10]$. We conducted this evaluation for both the context-less scenario and the scenario where previous titles were incorporated as context.  
\end{itemize}

The generation of section titles can be considered an extreme form of summarization. As such, we evaluate our models using established metrics in summarization: ROUGE \cite{lin_rouge_2004}, which measures the lexical overlap, and BARTScore \cite{yuan_bartscore_2021}, an increasingly used metric for semantic equivalence. The results of our experiments are shown in Table \ref{tab:title_gen}.

\paragraph{Importance of Context.} The results strongly underscore the difficulty in generating chapter titles solely based on the content of the current section without additional contextual information. An examination of the model's outputs (see Figure \ref{fig:titles_examples}) reveals how the lack of context leads to degraded coherence between titles of a document. Notably, the model has no knowledge about the placement of the current section in the document, leading to the repetition of functional titles such as \enquote{Intro}, which occur frequently in the dataset. Similarly, it also cannot generate sequential numbering or uphold uniform stylistic elements across the document's titles. In contrast, supplying the model with previously generated titles results in a meaningful increase in performance. This approach provides the model with the past context of the document structure, enabling stylistic continuity and a smoother flow between the titles.

\paragraph{Online Generation.} Expectedly, the model's performance improves as the input span $s$ increases. While diminishing returns are observable, they are less pronounced compared to the segmentation models. It is worth pointing out that the BART model with $s=3$ and which has access to the previous titles matches the performance of the BART model that has no context at all, once again underscoring the importance of context. Overall, considering both the segmentation models and the title generation models, we see 3 to 5 sentences as a reasonable trade-off between latency and performance for the future context size $c$ and the input span $s$. We note that $c$ can, in principle, be independently chosen from $s$ for practical smart chaptering systems. The overall latency for the title generation is dependent on the segmentation model, though, as segment boundaries for the span $s$ need to be determined before the generation of the title, as each sentence may belong to the next section.

\section{Related Work}

A number of benchmarks have been proposed for the text segmentation task. However, the majority are either small in size or synthetically constructed, often by concatenating documents or sections of documents \cite{choi_advances_2000,chen_global_2009,glavas_unsupervised_2016,glavas_training_2021}. Larger-scale benchmarks are scarce and limited to a narrow type of documents such as Wikipedia articles \cite{koshorek_text_2018} or specific domains like news \cite{liu_end--end_2022}\footnote{The news text segmentation dataset claimed to be made available by \citet{liu_end--end_2022} is presently inaccessible through the provided GitHub link, effectively yielding \textsc{Wiki-727K} to be the only large-scale benchmark.}. Equally, these datasets have in common that they only encompass documents that are structured in nature. In contrast, research on spoken, conversational, or generally unstructured or semi-structured content is still in its infancy. \citet{lv_vt-ssum_2021} and \citet{cho_toward_2022} separately proposed segmentation for lecture video transcripts. Although these works are related to our task, they are only confined to a single domain (lecture videos), while segments are artificially constructed based on the presentation slides. In the context of conversational content, \citet{ghosh_topic_2022} constructed a dataset by joining excerpts of different chat conversations, while \citet{zhong_qmsum_2021} provided a small set of meeting transcriptions segmented by topic shifts.

Regarding title generation, this study focuses on generating section headings and chapter titles for video transcripts. While research exists in title generation across various domains, including news headlines \cite{gu_generating_2020,liu_diverse_2020,cai_generating_2023}, product titles \cite{yang_multimodal_2023,zhu_dha_2022}, video titles \cite{zeng_title_2016,yu_tcr_2023}, StackOverflow posts \cite{liu_sotitle_2022,zhang_improving_2022,zhang_diverse_2023}, and pull requests \cite{irsan_autoprtitle_2022,zhang_automatic_2022}, our work addresses the unique challenges of video content structuring. One distinctive aspect lies in ensuring that all section headings of one document not only serve as informative signposts but also maintain a coherent and seamless flow between them. The closest work to ours has been conducted by \citet{zhang_outline_2019}, whose proposed model generates hierarchical outlines for Wikipedia documents. 

\section{Conclusion}
In this work, we present a novel benchmark for smart chaptering. The task aims to segment unstructured content, in particular speech, conversations, and transcriptions, in a linear sequence of chapters and provides each chapter with a title. We think this benchmark is a valuable addition to the text segmentation landscape as larger-scale, non-synthetic benchmarks are scarce, and previous research focused primarily on well-structured, homogeneous documents. As part of this, we propose an efficient and state-of-the-art hierarchical segmentation model and a corresponding title-generating model, both of which have also been architected to work online. 
By combining our proposed segmentation and title generation models, various practical applications are conceivable. For example, content creators, podcasters, and educators could use it to structure their content for their audience. We see our work also as a stepping stone to support even more unstructured content and speech in a broader scope, such as meetings.

\section*{Limitations}
Our study is subject to several limitations. First, the benchmark only provides English transcriptions which means it cannot assess text segmentation algorithms in languages other than English or be utilized in multilingual or cross-lingual contexts, an important area of research. Second, while the benchmark is inherently multi-modal, our evaluations were conducted solely on models trained on a single modality, which is the transcript, thus ignoring potentially valuable contextual information. Third, we want to note that the latency and real-timeliness of the online chaptering models depend on sentence lengths as the models operate on a sentence-level granularity. This dependence on sentence length restricts our ability to exert precise control over latency. Lastly, our title generation model suffers from exposure bias since it is trained using reference segmentations and prepending reference titles. In practical systems, we rely on both generated segment boundaries and titles, which can potentially lead to error propagation.

\section*{Acknowledgements}
This research is supported by the project "How is AI Changing Science? Research in the Era of Learning Algorithms" (HiAICS), funded by the Volkswagen Foundation. We also thank Jan Niehues for insightful discussions.

\bibliography{anthology,custom}

\begin{thebibliography}{43}
\expandafter\ifx\csname natexlab\endcsname\relax\def\natexlab#1{#1}\fi

\bibitem[{Beeferman et~al.(1999)Beeferman, Berger, and Lafferty}]{beeferman_statistical_1999}
Doug Beeferman, Adam Berger, and John Lafferty. 1999.
\newblock \href {https://doi.org/10.1023/A:1007506220214} {Statistical {Models} for {Text} {Segmentation}}.
\newblock \emph{Machine Learning}, 34(1):177--210.

\bibitem[{Bird et~al.(2009)Bird, Klein, and Loper}]{bird_natural_2009}
Steven Bird, Ewan Klein, and Edward Loper. 2009.
\newblock \emph{Natural language processing with {Python}}, 1st edition.
\newblock O'Reilly, Beijing ; Cambridge [Mass.].
\newblock OCLC: ocn301885973.

\bibitem[{Cai et~al.(2023)Cai, Song, Cho, Wang, Wang, Yu, Liu, and Yu}]{cai_generating_2023}
Pengshan Cai, Kaiqiang Song, Sangwoo Cho, Hongwei Wang, Xiaoyang Wang, Hong Yu, Fei Liu, and Dong Yu. 2023.
\newblock \href {https://doi.org/10.18653/v1/2023.acl-long.183} {Generating {User}-{Engaging} {News} {Headlines}}.
\newblock In \emph{Proceedings of the 61st {Annual} {Meeting} of the {Association} for {Computational} {Linguistics} ({Volume} 1: {Long} {Papers})}, pages 3265--3280, Toronto, Canada. Association for Computational Linguistics.

\bibitem[{Chen et~al.(2009)Chen, Branavan, Barzilay, and Karger}]{chen_global_2009}
Harr Chen, S.R.K. Branavan, Regina Barzilay, and David~R. Karger. 2009.
\newblock \href {https://aclanthology.org/N09-1042} {Global {Models} of {Document} {Structure} using {Latent} {Permutations}}.
\newblock In \emph{Proceedings of {Human} {Language} {Technologies}: {The} 2009 {Annual} {Conference} of the {North} {American} {Chapter} of the {Association} for {Computational} {Linguistics}}, pages 371--379, Boulder, Colorado. Association for Computational Linguistics.

\bibitem[{Chivers et~al.(2022)Chivers, P.~Jiang, Lee, Ng, Rapstine, and Storer}]{chivers_ants_2022}
Brian Chivers, Mason P.~Jiang, Wonhee Lee, Amy Ng, Natalya~I. Rapstine, and Alex Storer. 2022.
\newblock \href {https://doi.org/10.18653/v1/2022.deeplo-1.5} {{ANTS}: {A} {Framework} for {Retrieval} of {Text} {Segments} in {Unstructured} {Documents}}.
\newblock In \emph{Proceedings of the {Third} {Workshop} on {Deep} {Learning} for {Low}-{Resource} {Natural} {Language} {Processing}}, pages 38--47, Hybrid. Association for Computational Linguistics.

\bibitem[{Cho et~al.(2022)Cho, Song, Wang, Liu, and Yu}]{cho_toward_2022}
Sangwoo Cho, Kaiqiang Song, Xiaoyang Wang, Fei Liu, and Dong Yu. 2022.
\newblock \href {https://doi.org/10.18653/v1/2022.emnlp-main.8} {Toward {Unifying} {Text} {Segmentation} and {Long} {Document} {Summarization}}.
\newblock In \emph{Proceedings of the 2022 {Conference} on {Empirical} {Methods} in {Natural} {Language} {Processing}}, pages 106--118, Abu Dhabi, United Arab Emirates. Association for Computational Linguistics.

\bibitem[{Choi(2000)}]{choi_advances_2000}
Freddy Y.~Y. Choi. 2000.
\newblock \href {https://aclanthology.org/A00-2004} {Advances in domain independent linear text segmentation}.
\newblock In \emph{1st {Meeting} of the {North} {American} {Chapter} of the {Association} for {Computational} {Linguistics}}.

\bibitem[{Devlin et~al.(2019)Devlin, Chang, Lee, and Toutanova}]{devlin_bert_2019}
Jacob Devlin, Ming-Wei Chang, Kenton Lee, and Kristina Toutanova. 2019.
\newblock \href {https://doi.org/10.18653/v1/N19-1423} {{BERT}: {Pre}-training of {Deep} {Bidirectional} {Transformers} for {Language} {Understanding}}.
\newblock In \emph{Proceedings of the 2019 {Conference} of the {North} {American} {Chapter} of the {Association} for {Computational} {Linguistics}: {Human} {Language} {Technologies}, {Volume} 1 ({Long} and {Short} {Papers})}, pages 4171--4186, Minneapolis, Minnesota. Association for Computational Linguistics.

\bibitem[{Fournier(2013)}]{fournier_evaluating_2013}
Chris Fournier. 2013.
\newblock \href {https://aclanthology.org/P13-1167} {Evaluating {Text} {Segmentation} using {Boundary} {Edit} {Distance}}.
\newblock In \emph{Proceedings of the 51st {Annual} {Meeting} of the {Association} for {Computational} {Linguistics} ({Volume} 1: {Long} {Papers})}, pages 1702--1712, Sofia, Bulgaria. Association for Computational Linguistics.

\bibitem[{Ghosh et~al.(2022)Ghosh, Kajal, Kamath, Shrivastava, Basu, and Srinivasan}]{ghosh_topic_2022}
Reshmi Ghosh, Harjeet~Singh Kajal, Sharanya Kamath, Dhuri Shrivastava, Samyadeep Basu, and Soundararajan Srinivasan. 2022.
\newblock \href {https://doi.org/10.48550/arXiv.2211.14954} {Topic {Segmentation} in the {Wild}: {Towards} {Segmentation} of {Semi}-structured \& {Unstructured} {Chats}}.
\newblock ArXiv:2211.14954 [cs].

\bibitem[{Glavaš et~al.(2021)Glavaš, Ganesh, and Somasundaran}]{glavas_training_2021}
Goran Glavaš, Ananya Ganesh, and Swapna Somasundaran. 2021.
\newblock \href {https://aclanthology.org/2021.bea-1.11} {Training and {Domain} {Adaptation} for {Supervised} {Text} {Segmentation}}.
\newblock In \emph{Proceedings of the 16th {Workshop} on {Innovative} {Use} of {NLP} for {Building} {Educational} {Applications}}, pages 110--116, Online. Association for Computational Linguistics.

\bibitem[{Glavaš et~al.(2016)Glavaš, Nanni, and Ponzetto}]{glavas_unsupervised_2016}
Goran Glavaš, Federico Nanni, and Simone~Paolo Ponzetto. 2016.
\newblock \href {https://doi.org/10.18653/v1/S16-2016} {Unsupervised {Text} {Segmentation} {Using} {Semantic} {Relatedness} {Graphs}}.
\newblock In \emph{Proceedings of the {Fifth} {Joint} {Conference} on {Lexical} and {Computational} {Semantics}}, pages 125--130, Berlin, Germany. Association for Computational Linguistics.

\bibitem[{Gu et~al.(2020)Gu, Mao, Han, Liu, Wu, Yu, Finnie, Yu, Zhai, and Zukoski}]{gu_generating_2020}
Xiaotao Gu, Yuning Mao, Jiawei Han, Jialu Liu, You Wu, Cong Yu, Daniel Finnie, Hongkun Yu, Jiaqi Zhai, and Nicholas Zukoski. 2020.
\newblock \href {https://doi.org/10.1145/3366423.3380247} {Generating {Representative} {Headlines} for {News} {Stories}}.
\newblock In \emph{Proceedings of {The} {Web} {Conference} 2020}, {WWW} '20, pages 1773--1784, New York, NY, USA. Association for Computing Machinery.

\bibitem[{Irsan et~al.(2022)Irsan, Zhang, Thung, Lo, and Jiang}]{irsan_autoprtitle_2022}
Ivana~Clairine Irsan, Ting Zhang, Ferdian Thung, David Lo, and Lingxiao Jiang. 2022.
\newblock \href {https://doi.org/10.1109/ICSME55016.2022.00058} {{AutoPRTitle}: {A} {Tool} for {Automatic} {Pull} {Request} {Title} {Generation}}.
\newblock pages 454--458. IEEE Computer Society.

\bibitem[{Koshorek et~al.(2018)Koshorek, Cohen, Mor, Rotman, and Berant}]{koshorek_text_2018}
Omri Koshorek, Adir Cohen, Noam Mor, Michael Rotman, and Jonathan Berant. 2018.
\newblock \href {https://doi.org/10.18653/v1/N18-2075} {Text {Segmentation} as a {Supervised} {Learning} {Task}}.
\newblock In \emph{Proceedings of the 2018 {Conference} of the {North} {American} {Chapter} of the {Association} for {Computational} {Linguistics}: {Human} {Language} {Technologies}, {Volume} 2 ({Short} {Papers})}, pages 469--473, New Orleans, Louisiana. Association for Computational Linguistics.

\bibitem[{Lewis et~al.(2020)Lewis, Liu, Goyal, Ghazvininejad, Mohamed, Levy, Stoyanov, and Zettlemoyer}]{lewis_bart_2020}
Mike Lewis, Yinhan Liu, Naman Goyal, Marjan Ghazvininejad, Abdelrahman Mohamed, Omer Levy, Veselin Stoyanov, and Luke Zettlemoyer. 2020.
\newblock \href {https://doi.org/10.18653/v1/2020.acl-main.703} {{BART}: {Denoising} {Sequence}-to-{Sequence} {Pre}-training for {Natural} {Language} {Generation}, {Translation}, and {Comprehension}}.
\newblock In \emph{Proceedings of the 58th {Annual} {Meeting} of the {Association} for {Computational} {Linguistics}}, pages 7871--7880, Online. Association for Computational Linguistics.

\bibitem[{Lin(2004)}]{lin_rouge_2004}
Chin-Yew Lin. 2004.
\newblock \href {https://aclanthology.org/W04-1013} {{ROUGE}: {A} {Package} for {Automatic} {Evaluation} of {Summaries}}.
\newblock In \emph{Text {Summarization} {Branches} {Out}}, pages 74--81, Barcelona, Spain. Association for Computational Linguistics.

\bibitem[{Liu et~al.(2020)Liu, Gong, Yan, Fu, Shao, Jiang, Lv, and Duan}]{liu_diverse_2020}
Dayiheng Liu, Yeyun Gong, Yu~Yan, Jie Fu, Bo~Shao, Daxin Jiang, Jiancheng Lv, and Nan Duan. 2020.
\newblock \href {https://doi.org/10.18653/v1/2020.emnlp-main.505} {Diverse, {Controllable}, and {Keyphrase}-{Aware}: {A} {Corpus} and {Method} for {News} {Multi}-{Headline} {Generation}}.
\newblock In \emph{Proceedings of the 2020 {Conference} on {Empirical} {Methods} in {Natural} {Language} {Processing} ({EMNLP})}, pages 6241--6250, Online. Association for Computational Linguistics.

\bibitem[{Liu et~al.(2022{\natexlab{a}})Liu, Yang, Chen, and Yu}]{liu_sotitle_2022}
Ke~Liu, Guang Yang, Xiang Chen, and Chi Yu. 2022{\natexlab{a}}.
\newblock \href {https://doi.org/10.1109/SANER53432.2022.00075} {{SOTitle}: {A} {Transformer}-based {Post} {Title} {Generation} {Approach} for {Stack} {Overflow}}.
\newblock pages 577--588. IEEE Computer Society.
\newblock ISSN: 1534-5351.

\bibitem[{Liu et~al.(2022{\natexlab{b}})Liu, Zhu, and Zeng}]{liu_end--end_2022}
Yang Liu, Chenguang Zhu, and Michael Zeng. 2022{\natexlab{b}}.
\newblock \href {https://doi.org/10.18653/v1/2022.findings-acl.46} {End-to-{End} {Segmentation}-based {News} {Summarization}}.
\newblock In \emph{Findings of the {Association} for {Computational} {Linguistics}: {ACL} 2022}, pages 544--554, Dublin, Ireland. Association for Computational Linguistics.

\bibitem[{Liu et~al.(2019)Liu, Ott, Goyal, Du, Joshi, Chen, Levy, Lewis, Zettlemoyer, and Stoyanov}]{liu_roberta_2019}
Yinhan Liu, Myle Ott, Naman Goyal, Jingfei Du, Mandar Joshi, Danqi Chen, Omer Levy, Mike Lewis, Luke Zettlemoyer, and Veselin Stoyanov. 2019.
\newblock \href {https://doi.org/10.48550/arXiv.1907.11692} {{RoBERTa}: {A} {Robustly} {Optimized} {BERT} {Pretraining} {Approach}}.
\newblock ArXiv:1907.11692 [cs].

\bibitem[{Lukasik et~al.(2020)Lukasik, Dadachev, Papineni, and Simões}]{lukasik_text_2020}
Michal Lukasik, Boris Dadachev, Kishore Papineni, and Gonçalo Simões. 2020.
\newblock \href {https://doi.org/10.18653/v1/2020.emnlp-main.380} {Text {Segmentation} by {Cross} {Segment} {Attention}}.
\newblock In \emph{Proceedings of the 2020 {Conference} on {Empirical} {Methods} in {Natural} {Language} {Processing} ({EMNLP})}, pages 4707--4716, Online. Association for Computational Linguistics.

\bibitem[{Lv et~al.(2021)Lv, Cui, Vasilijevic, and Wei}]{lv_vt-ssum_2021}
Tengchao Lv, Lei Cui, Momcilo Vasilijevic, and Furu Wei. 2021.
\newblock \href {https://doi.org/10.48550/arXiv.2106.05606} {{VT}-{SSum}: {A} {Benchmark} {Dataset} for {Video} {Transcript} {Segmentation} and {Summarization}}.
\newblock ArXiv:2106.05606 [cs].

\bibitem[{McInnes et~al.(2018)McInnes, Healy, Saul, and Großberger}]{mcinnes_umap_2018}
Leland McInnes, John Healy, Nathaniel Saul, and Lukas Großberger. 2018.
\newblock \href {https://doi.org/10.21105/joss.00861} {{UMAP}: {Uniform} {Manifold} {Approximation} and {Projection}}.
\newblock \emph{Journal of Open Source Software}, 3(29):861.

\bibitem[{Prince and Labadié(2007)}]{prince_text_2007}
Violaine Prince and Alexandre Labadié. 2007.
\newblock \href {https://doi.org/10.1007/978-3-540-73351-5_26} {Text {Segmentation} {Based} on {Document} {Understanding} for {Information} {Retrieval}}.
\newblock In \emph{Natural {Language} {Processing} and {Information} {Systems}}, Lecture {Notes} in {Computer} {Science}, pages 295--304, Berlin, Heidelberg. Springer.

\bibitem[{Reimers and Gurevych(2019)}]{reimers_sentence-bert_2019}
Nils Reimers and Iryna Gurevych. 2019.
\newblock \href {https://doi.org/10.18653/v1/D19-1410} {Sentence-{BERT}: {Sentence} {Embeddings} using {Siamese} {BERT}-{Networks}}.
\newblock In \emph{Proceedings of the 2019 {Conference} on {Empirical} {Methods} in {Natural} {Language} {Processing} and the 9th {International} {Joint} {Conference} on {Natural} {Language} {Processing} ({EMNLP}-{IJCNLP})}, pages 3982--3992, Hong Kong, China. Association for Computational Linguistics.

\bibitem[{Retkowski(2023)}]{sudmann_current_2023}
Fabian Retkowski. 2023.
\newblock \href {https://doi.org/10.14361/9783839467664-016} {The current state of summarization}.
\newblock In Andreas Sudmann, Anna Echterhölter, Markus Ramsauer, Fabian Retkowski, Jens Schröter, and Alexander Waibel, editors, \emph{Beyond {Quantity}. {Research} with {Subsymbolic} {AI}}, 1 edition, pages 291--312. transcript Verlag, Bielefeld, Germany.

\bibitem[{Shtekh et~al.(2018)Shtekh, Kazakova, Nikitinsky, and Skachkov}]{shtekh_applying_2018}
Gennady Shtekh, Polina Kazakova, Nikita Nikitinsky, and Nikolay Skachkov. 2018.
\newblock \href {https://doi.org/10.1145/3290621.3290630} {Applying {Topic} {Segmentation} to {Document}-{Level} {Information} {Retrieval}}.
\newblock In \emph{Proceedings of the 14th {Central} and {Eastern} {European} {Software} {Engineering} {Conference} {Russia}}, {CEE}-{SECR} '18, pages 1--6, New York, NY, USA. Association for Computing Machinery.

\bibitem[{Su et~al.(2023)Su, Shi, Kasai, Wang, Hu, Ostendorf, Yih, Smith, Zettlemoyer, and Yu}]{su_one_2023}
Hongjin Su, Weijia Shi, Jungo Kasai, Yizhong Wang, Yushi Hu, Mari Ostendorf, Wen-tau Yih, Noah~A. Smith, Luke Zettlemoyer, and Tao Yu. 2023.
\newblock \href {https://doi.org/10.18653/v1/2023.findings-acl.71} {One {Embedder}, {Any} {Task}: {Instruction}-{Finetuned} {Text} {Embeddings}}.
\newblock In \emph{Findings of the {Association} for {Computational} {Linguistics}: {ACL} 2023}, pages 1102--1121, Toronto, Canada. Association for Computational Linguistics.

\bibitem[{Su et~al.(2021)Su, Lu, Pan, Murtadha, Wen, and Liu}]{su_roformer_2021}
Jianlin Su, Yu~Lu, Shengfeng Pan, Ahmed Murtadha, Bo~Wen, and Yunfeng Liu. 2021.
\newblock \href {https://arxiv.org/abs/2104.09864v4} {{RoFormer}: {Enhanced} {Transformer} with {Rotary} {Position} {Embedding}}.

\bibitem[{Touvron et~al.(2023)Touvron, Martin, Stone, Albert, Almahairi, Babaei, Bashlykov, Batra, Bhargava, Bhosale, Bikel, Blecher, Ferrer, Chen, Cucurull, Esiobu, Fernandes, Fu, Fu, Fuller, Gao, Goswami, Goyal, Hartshorn, Hosseini, Hou, Inan, Kardas, Kerkez, Khabsa, Kloumann, Korenev, Koura, Lachaux, Lavril, Lee, Liskovich, Lu, Mao, Martinet, Mihaylov, Mishra, Molybog, Nie, Poulton, Reizenstein, Rungta, Saladi, Schelten, Silva, Smith, Subramanian, Tan, Tang, Taylor, Williams, Kuan, Xu, Yan, Zarov, Zhang, Fan, Kambadur, Narang, Rodriguez, Stojnic, Edunov, and Scialom}]{touvron_llama_2023}
Hugo Touvron, Louis Martin, Kevin Stone, Peter Albert, Amjad Almahairi, Yasmine Babaei, Nikolay Bashlykov, Soumya Batra, Prajjwal Bhargava, Shruti Bhosale, Dan Bikel, Lukas Blecher, Cristian~Canton Ferrer, Moya Chen, Guillem Cucurull, David Esiobu, Jude Fernandes, Jeremy Fu, Wenyin Fu, Brian Fuller, Cynthia Gao, Vedanuj Goswami, Naman Goyal, Anthony Hartshorn, Saghar Hosseini, Rui Hou, Hakan Inan, Marcin Kardas, Viktor Kerkez, Madian Khabsa, Isabel Kloumann, Artem Korenev, Punit~Singh Koura, Marie-Anne Lachaux, Thibaut Lavril, Jenya Lee, Diana Liskovich, Yinghai Lu, Yuning Mao, Xavier Martinet, Todor Mihaylov, Pushkar Mishra, Igor Molybog, Yixin Nie, Andrew Poulton, Jeremy Reizenstein, Rashi Rungta, Kalyan Saladi, Alan Schelten, Ruan Silva, Eric~Michael Smith, Ranjan Subramanian, Xiaoqing~Ellen Tan, Binh Tang, Ross Taylor, Adina Williams, Jian~Xiang Kuan, Puxin Xu, Zheng Yan, Iliyan Zarov, Yuchen Zhang, Angela Fan, Melanie Kambadur, Sharan Narang, Aurelien Rodriguez, Robert Stojnic, Sergey Edunov, and Thomas
  Scialom. 2023.
\newblock \href {https://doi.org/10.48550/arXiv.2307.09288} {Llama 2: {Open} {Foundation} and {Fine}-{Tuned} {Chat} {Models}}.
\newblock ArXiv:2307.09288 [cs].

\bibitem[{Wang et~al.(2020)Wang, Wei, Dong, Bao, Yang, and Zhou}]{wang_minilm_2020}
Wenhui Wang, Furu Wei, Li~Dong, Hangbo Bao, Nan Yang, and Ming Zhou. 2020.
\newblock \href {https://proceedings.neurips.cc/paper_files/paper/2020/file/3f5ee243547dee91fbd053c1c4a845aa-Paper.pdf} {{MiniLM}: {Deep} {Self}-{Attention} {Distillation} for {Task}-{Agnostic} {Compression} of {Pre}-{Trained} {Transformers}}.
\newblock In \emph{Advances in {Neural} {Information} {Processing} {Systems}}, volume~33, pages 5776--5788. Curran Associates, Inc.

\bibitem[{Yang et~al.(2023)Yang, Liu, Li, Yin, You, Yin, and Zou}]{yang_multimodal_2023}
Bang Yang, Fenglin Liu, Zheng Li, Qingyu Yin, Chenyu You, Bing Yin, and Yuexian Zou. 2023.
\newblock \href {https://doi.org/10.18653/v1/2023.findings-acl.166} {Multimodal {Prompt} {Learning} for {Product} {Title} {Generation} with {Extremely} {Limited} {Labels}}.
\newblock In \emph{Findings of the {Association} for {Computational} {Linguistics}: {ACL} 2023}, pages 2652--2665, Toronto, Canada. Association for Computational Linguistics.

\bibitem[{Yu et~al.(2023)Yu, Yang, Guo, Liu, Xu, and Niu}]{yu_tcr_2023}
Yakun Yu, Jiuding Yang, Weidong Guo, Hui Liu, Yu~Xu, and Di~Niu. 2023.
\newblock \href {https://doi.org/10.1007/978-3-031-33380-4_19} {{TCR}: {Short} {Video} {Title} {Generation} and {Cover} {Selection} with {Attention} {Refinement}}.
\newblock In \emph{Advances in {Knowledge} {Discovery} and {Data} {Mining}}, Lecture {Notes} in {Computer} {Science}, pages 245--256, Cham. Springer Nature Switzerland.

\bibitem[{Yuan et~al.(2021)Yuan, Neubig, and Liu}]{yuan_bartscore_2021}
Weizhe Yuan, Graham Neubig, and Pengfei Liu. 2021.
\newblock \href {https://proceedings.neurips.cc/paper_files/paper/2021/file/e4d2b6e6fdeca3e60e0f1a62fee3d9dd-Paper.pdf} {{BARTScore}: {Evaluating} {Generated} {Text} as {Text} {Generation}}.
\newblock In \emph{Advances in {Neural} {Information} {Processing} {Systems}}, volume~34, pages 27263--27277. Curran Associates, Inc.

\bibitem[{Zechner and Waibel(2000)}]{zechner_diasumm_2000}
Klaus Zechner and Alex Waibel. 2000.
\newblock \href {https://aclanthology.org/C00-2140} {{DIASUMM}: {Flexible} {Summarization} of {Spontaneous} {Dialogues} in {Unrestricted} {Domains}}.
\newblock In \emph{{COLING} 2000 {Volume} 2: {The} 18th {International} {Conference} on {Computational} {Linguistics}}.

\bibitem[{Zeng et~al.(2016)Zeng, Chen, Niebles, and Sun}]{zeng_title_2016}
Kuo-Hao Zeng, Tseng-Hung Chen, Juan~Carlos Niebles, and Min Sun. 2016.
\newblock \href {https://doi.org/10.1007/978-3-319-46475-6_38} {Title {Generation} for {User} {Generated} {Videos}}.
\newblock In \emph{Computer {Vision} – {ECCV} 2016}, Lecture {Notes} in {Computer} {Science}, pages 609--625, Cham. Springer International Publishing.

\bibitem[{Zhang et~al.(2023)Zhang, Liu, Wan, Yu, Liu, and Keung}]{zhang_diverse_2023}
Fengji Zhang, Jin Liu, Yao Wan, Xiao Yu, Xiao Liu, and Jacky Keung. 2023.
\newblock \href {https://doi.org/10.1016/j.jss.2023.111672} {Diverse title generation for {Stack} {Overflow} posts with multiple-sampling-enhanced transformer}.
\newblock \emph{Journal of Systems and Software}, 200:111672.

\bibitem[{Zhang et~al.(2022{\natexlab{a}})Zhang, Yu, Keung, Li, Xie, Yang, Ma, and Zhang}]{zhang_improving_2022}
Fengji Zhang, Xiao Yu, Jacky Keung, Fuyang Li, Zhiwen Xie, Zhen Yang, Caoyuan Ma, and Zhimin Zhang. 2022{\natexlab{a}}.
\newblock \href {https://doi.org/10.1016/j.infsof.2022.106922} {Improving {Stack} {Overflow} question title generation with copying enhanced {CodeBERT} model and bi-modal information}.
\newblock \emph{Information and Software Technology}, 148:106922.

\bibitem[{Zhang et~al.(2019)Zhang, Guo, Fan, Lan, and Cheng}]{zhang_outline_2019}
Ruqing Zhang, Jiafeng Guo, Yixing Fan, Yanyan Lan, and Xueqi Cheng. 2019.
\newblock \href {https://doi.org/10.1145/3331184.3331208} {Outline {Generation}: {Understanding} the {Inherent} {Content} {Structure} of {Documents}}.
\newblock In \emph{Proceedings of the 42nd {International} {ACM} {SIGIR} {Conference} on {Research} and {Development} in {Information} {Retrieval}}, {SIGIR}'19, pages 745--754, New York, NY, USA. Association for Computing Machinery.

\bibitem[{Zhang et~al.(2022{\natexlab{b}})Zhang, Irsan, Thung, Han, Lo, and Jiang}]{zhang_automatic_2022}
Ting Zhang, Ivana~Clairine Irsan, Ferdian Thung, DongGyun Han, David Lo, and Lingxiao Jiang. 2022{\natexlab{b}}.
\newblock \href {https://doi.org/10.1109/ICSME55016.2022.00015} {Automatic {Pull} {Request} {Title} {Generation}}.
\newblock In \emph{2022 {IEEE} {International} {Conference} on {Software} {Maintenance} and {Evolution} ({ICSME})}, pages 71--81.
\newblock ISSN: 2576-3148.

\bibitem[{Zhong et~al.(2021)Zhong, Yin, Yu, Zaidi, Mutuma, Jha, Awadallah, Celikyilmaz, Liu, Qiu, and Radev}]{zhong_qmsum_2021}
Ming Zhong, Da~Yin, Tao Yu, Ahmad Zaidi, Mutethia Mutuma, Rahul Jha, Ahmed~Hassan Awadallah, Asli Celikyilmaz, Yang Liu, Xipeng Qiu, and Dragomir Radev. 2021.
\newblock \href {https://doi.org/10.18653/v1/2021.naacl-main.472} {{QMSum}: {A} {New} {Benchmark} for {Query}-based {Multi}-domain {Meeting} {Summarization}}.
\newblock In \emph{Proceedings of the 2021 {Conference} of the {North} {American} {Chapter} of the {Association} for {Computational} {Linguistics}: {Human} {Language} {Technologies}}, pages 5905--5921, Online. Association for Computational Linguistics.

\bibitem[{Zhu et~al.(2022)Zhu, Zhang, Zhang, Zhou, Feng, Wu, Da, and Zeng}]{zhu_dha_2022}
Wenya Zhu, Yinghua Zhang, Yu~Zhang, Yuhang Zhou, Yinfu Feng, Yuxiang Wu, Qing Da, and Anxiang Zeng. 2022.
\newblock \href {https://doi.org/10.1007/978-3-031-05981-0_22} {{DHA}: {Product} {Title} {Generation} with {Discriminative} {Hierarchical} {Attention} for {E}-commerce}.
\newblock In \emph{Advances in {Knowledge} {Discovery} and {Data} {Mining}}, Lecture {Notes} in {Computer} {Science}, pages 275--287, Cham. Springer International Publishing.

\end{thebibliography}
\bibliographystyle{acl_natbib}

\appendix

\setcounter{table}{0}
\renewcommand{\thetable}{A\arabic{table}}

\setcounter{figure}{0}
\renewcommand{\thefigure}{A\arabic{figure}}

\onecolumn

\section{YouTube Chapters}
\begin{figure}[h]
\centering
\includegraphics[width=\textwidth,trim={0 0 0 0.5cm},clip]{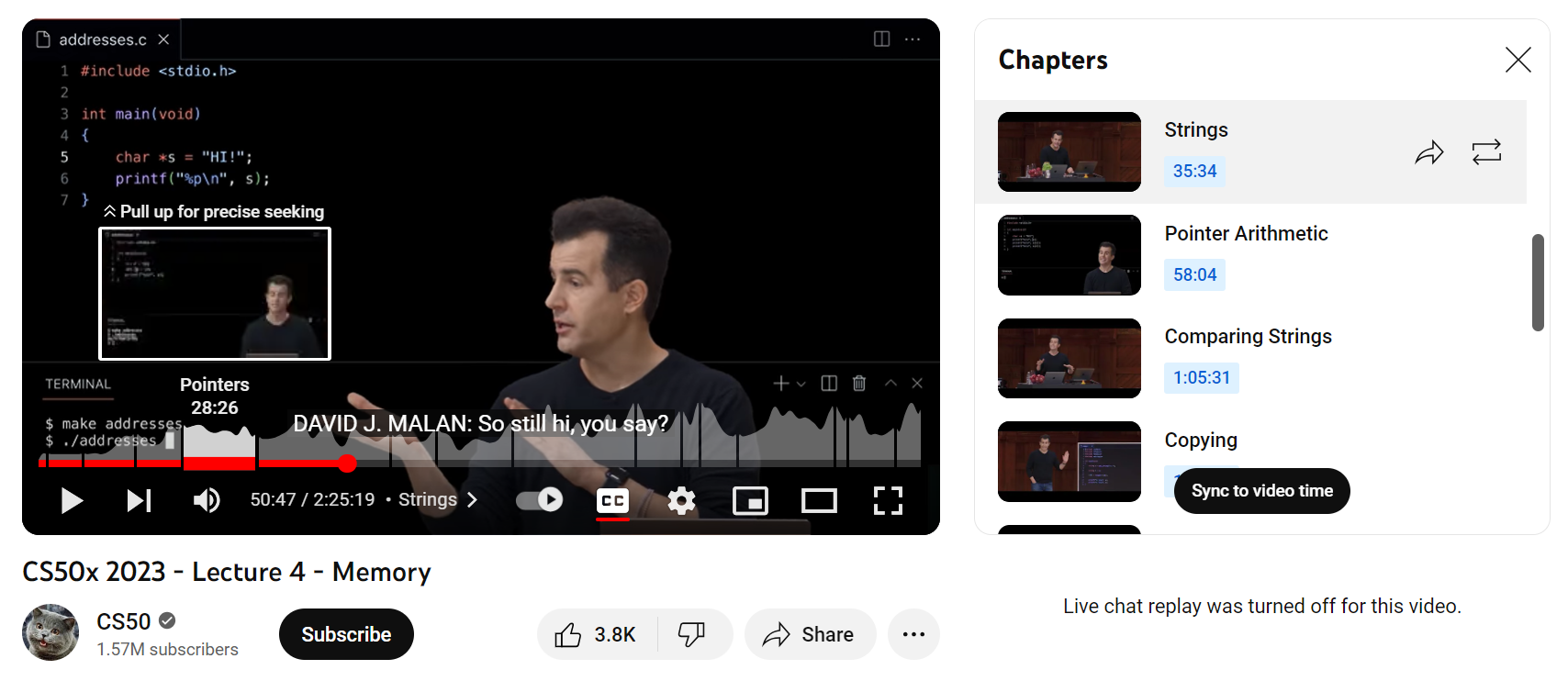}
\caption{A screenshot of a YouTube video featuring segments as chapters, which form the basis of our new text segmentation benchmark \textsc{YTSeg}.}
\label{fig:screenshot}
\end{figure}

\section{Data Splits}

\begin{table}[H]
    \centering
    \begin{subtable}[t]{0.4\textwidth}
        \centering
        \begin{tabular}{>{\raggedright\arraybackslash}p{2cm}S[table-alignment=right,
                  round-mode=places,
                  round-precision=1,
                  table-format=5.0,
                  zero-decimal-to-integer,
                  input-symbols={()\%},,
                  table-space-text-post={\enspace(11.1\,\%)},
                  table-align-text-post=true]}
            \toprule
            \textbf{Partition} & \textbf{\# Examples} \\
            \midrule
            Training & 16404 \mypc{85} \\
            Validation & 1447 \mypc{7.5} \\
            Testing & 1448 \mypc{7.5} \\
            \midrule
            Total & 19229 \\
            \bottomrule
        \end{tabular}
        \caption{\textsc{YTSeg} data split}
        \label{table:ytseg_split}
    \end{subtable}
    \quad
    \begin{subtable}[t]{0.4\textwidth}
        \centering
        \begin{tabular}{>{\raggedright\arraybackslash}p{2cm}S[table-alignment=right,
                  round-mode=places,
                  round-precision=1,
                  table-format=5.0,
                  zero-decimal-to-integer,
                  input-symbols={()\%},,
                  table-space-text-post={\enspace(11.1\,\%)},
                  table-align-text-post=true]}
            \toprule
            \textbf{Partition} & \textbf{\# Examples} \\
            \midrule
            Training & 146907 \mypc{84.8} \\
            Validation & 13206 \mypc{7.6} \\
            Testing & 13082 \mypc{7.6} \\
            \midrule
            Total & 173195 \\
            \bottomrule
        \end{tabular}
        \caption{\textsc{YTSeg[Titles]} data split}
        \label{table:ytsegtitle_split}
    \end{subtable}
    \caption{Data splits for \textsc{YTSeg} and \textsc{YTSeg[Titles]}}
    \end{table}

\section{Hyperparameters}

We manually tuned hyperparameters and provide the parameter sets responsible for the results disclosed in this research. While training, we continuously calculated the relevant test metrics (such as the F1 score for segmentation) on the validation data and performed model selection based on this information.

\begin{table}[H]
    \centering
    \begin{tabular}{>{\raggedright\arraybackslash}p{3.75cm}r}
        \toprule
        \textbf{Hyperparameter} & \textbf{Value} \\
        \midrule
        Sentence Encoder & \texttt{\small \href{https://huggingface.co/sentence-transformers/all-MiniLM-L6-v2}{sentence-transformers/all-MiniLM-L6-v2}} \\ 
        Loss Function & Weighted Binary Cross-Entropy \\
        Cross-Entropy Weights & $[1,2]$ \\
        Learning Rate & $2.5 \times 10^{-5}$ \\
        Batch Size & 115,000 Tokens \\
        Epochs & 15 \\
        Learning Rate Schedule & Cosine \\
        Optimizer & AdamW \\
        Dropout Rate & 0.1 \\
        Gradient Sampling Rate & 0.5 \\
        \bottomrule
    \end{tabular}
    \caption{Hyperparameters for MiniSeg training on the \textsc{YTSeg} dataset}
    \label{tab:hyperparameters}
\end{table}

\begin{multicols}{2}

\begin{table}[H]
    \centering
    \begin{tabular}{>{\raggedright\arraybackslash}rrr}
        \toprule
        $\boldsymbol{c}$ & 
        $\boldsymbol{M}$ & 
        $\boldsymbol{\alpha}$ \\
        \midrule
        1 & 1 & $[1]$ \\ 
        3 & 2 & $[2,1]$ \\
        5 & 3 & $[2,2,1]$ \\
        8 & 4 & $[2,2,2,2]$ \\
        10 & 5 & $[2,2,2,2,2]$ \\
        20 & 7 & $[4,4,4,2,2,2,2]$ \\
        \bottomrule
    \end{tabular}
    \caption{Overview of the partitioning $\alpha$ and the number of future-context-accumulating layers $M$ used in the corresponding online segmentation models with future context size $c$.}
    \label{tab:partition}
\end{table}

\begin{table}[H]
    \centering
    \tabcolsep=0.09cm
    \begin{tabular}{>{\raggedright\arraybackslash}p{3.75cm}r}
        \toprule
        \textbf{Hyperparameter} & \textbf{Value} \\
        \midrule
        Base & \texttt{\small \href{https://huggingface.co/facebook/bart-large}{facebook/bart-large}} \\
        Learning Rate & $5 \times 10^{-5}$ \\
        Batch Size & 10,000 Tokens \\
        Epochs & 2 \\
        Learning Rate Schedule & Cosine \\
        Optimizer & AdamW \\
        Dropout Rate & 0.1 \\
        \cdashlinelr{1-2}
        Decoding Strategy & Beam Sampling \\
        Beam Size & $5$ \\
        Top $k$ & $50$ \\
        Top $p$ & $0.95$ \\
        \bottomrule
    \end{tabular}
    \caption{Hyperparameters for training and evaluating the title generation model on \textsc{YTSeg[Titles]}}
    \label{tab:hyperparameters}
\end{table}
\end{multicols}

\section{Evaluation}
We use the \texttt{segeval}\footnote{\url{https://segeval.readthedocs.io/}} package \cite{fournier_evaluating_2013} for the computation of segmentation performance metrics, including $P_k$ and Boundary Similarity. In both cases, we adhere to the default parameter settings. For the evaluation of the title generation models, we rely on the \texttt{rouge-metric}\footnote{\url{https://github.com/li-plus/rouge-metric}} package that wraps and reimplements the official ROUGE-1.5.5 Perl script \cite{lin_rouge_2004}. Lastly, for the BARTScore, we utilize the official implementation and ParaBank2-trained BART model\footnote{\url{https://github.com/neulab/BARTScore}} provided by \citet{yuan_bartscore_2021}.

\section{Seed Keywords}
\label{sec:seed}
\begin{multicols}{4}
\begin{itemize}
\item lecture
\item podcast
\item meetup
\item theory
\item math
\item physics
\item chemistry
\item climate history
\item geometry
\item electrical engineering
\item media theory
\item fashion
\item tech
\item explained
\item "analysis of"
\item "introducing"
\item "simplified"
\item "explanation of"
\item "the art of"
\item "mechanics of"
\item "recent advances in"
\item "in a nutshell"
\item "the theory of"
\item "guide to"
\item "why do *"
\item "why does *"
\item "exploring *"
\item "* talk"
\item robotics
\item computer vision
\item virtual reality
\item insurance
\item dietary
\item azure
\item brain
\item linear algebra
\item oled
\item silicon
\item linux
\item deployment
\item nature
\item adobe
\item ui design
\item rna
\item pytorch
\item self driving cars
\item machine learning
\item data science
\end{itemize}
\end{multicols}
\end{document}